\documentclass[letterpaper]{article}
\usepackage[letterpaper]{geometry}

\usepackage[utf8]{inputenc}
\usepackage[T1]{fontenc} 
\usepackage[cjk]{kotex}

\usepackage{amsmath}
\usepackage{graphicx}
\usepackage{url}
\usepackage{xcolor}
\usepackage{latexsym}

\usepackage{hyperref}
\definecolor{darkblue}{rgb}{0, 0, 0.5}
\hypersetup{colorlinks=true,citecolor=darkblue, linkcolor=darkblue, urlcolor=darkblue}
\usepackage{tabularx}
\usepackage{datetime}
\usepackage{arydshln}

\usepackage{setspace}

\usepackage[authoryear,round,longnamesfirst]{natbib}
\usepackage[ruled,vlined]{algorithm2e}
\usepackage[linguistics]{forest} 
\usepackage{synttree}
\usepackage{subcaption}
\usepackage{tikz-dependency}

\usepackage{gb4e}

\title{Word segmentation granularity in Korean}

\author{ 
Jungyeul Park$^{1}$\thanks{Corresponding author} \qquad Mija Kim$^{2}$ \\
{\small $^{1}$Department of Linguistics, The University of British Columbia} \\
{\small Vancouver, BC. V6T 1Z4, Canada.  
\texttt{jungyeul@mail.ubc.ca}}\\
{\small $^{2}$Department of English Language and Literature, Kyung Hee University} \\
{\small Seoul, 02447, South Korea. \texttt{3mjkim@khu.ac.kr}}
}
\date{}

\begin{document}

\maketitle

\doublespacing

\begin{abstract}
This paper describes word {segmentation} granularity in Korean language processing. From a word separated by blank space, which is termed an eojeol, to a sequence of morphemes in Korean, there are multiple possible levels of word segmentation granularity in Korean. For specific language processing and corpus annotation tasks, several different granularity levels have been proposed and utilized, because the agglutinative languages including Korean language have a one-to-one mapping between functional morpheme and syntactic category. Thus, we analyze these different granularity levels, presenting the examples of Korean language processing systems for future reference. Interestingly, the granularity by separating only functional morphemes including case markers and verbal endings, and keeping other suffixes for morphological derivation results in the optimal performance for phrase structure parsing. This contradicts previous best practices for Korean language processing, which has been the de facto standard for various applications that require separating all morphemes.

\noindent\textbf{keywords}: {word {segmentation} granularity , morphological segmentation, agglutinative language, evaluation }
\end{abstract}


\section{Introduction}

Morphological analysis for Korean has been based on an eojeol, which has been considered as a basic segmentation unit in Korean delimited by white blank spaces in a sentence. Almost all of the language processing systems and language data sets previously developed for Korean have utilized this eojeol as a fundamental unit of analysis. Given that Korean is an agglutinative language, joining content and functional morphemes of words is very productive and the number of their combinations is exponential. We can treat a given noun or verb as a stem (also content) followed by several functional morphemes in Korean. Some of these morphemes can, sometimes, be assigned its syntactic category. Let us consider the sentence in~\eqref{ungaro}.

The corresponding morphological analysis is also provided in Figure~\ref{ungaro-pos}. 
\textit{Unggaro} (`Ungaro') is a content morpheme (a proper noun) and a postposition \textit{-ga} (nominative) is a functional morpheme. They form together a single eojeol (or word) \textit{unggaro-ga} (`Ungaro+\textsc{nom}'). 
For the sake of convenience, we add a figure dash (\textit{-}) at the beginning of functional morphemes, such as \textit{-ga} (\textsc{nom}) to distinguish between content and functional morphemes. The nominative case markers \textit{-ga} or \textit{-i} may vary depending on the previous letter --- vowel or consonant. A predicate \textit{naseo-eoss-da} also consists of the content morpheme \textit{naseo} (`become') and its functional morphemes, \textit{-eoss} (`\textsc{past}') and \textit{-da} (`\textsc{decl}'), respectively.

\begin{exe}
 \ex \label{ungaro}
 \gll
 \textit{peurangseu-ui} {\textit{segye-jeok-i-n}} \textit{uisang} \textit{dijaineo} \textit{emmanuel} {\textit{unggaro-ga}} \textit{silnae} \textit{jangsik-yong   } \textit{jikmul   } \textit{dijaineo-ro} \textit{{naseo-eoss-da}.}\\ 
 France-\textsc{gen} world~class-\textsc{rel} fashion designer Emanuel Ungaro-\textsc{nom} interior    decoration textile designer-\textsc{ajt} become-\textsc{past}-\textsc{decl}\\
 \trans 'The world-class French fashion designer Emanuel Ungaro became an interior textile designer.' 
\end{exe}

\begin{figure}
\begin{center}
{\small
\begin{tabular} {rlcl l}
프랑스의 & \textit{peurangseu-ui} & &\textit{peurangseu}/NNP+\textit{ui}/JKG & France-\textsc{gen}\\
세계적인 &\textit{segye-jeok-i-n} && \textit{segye}/NNG+\textit{jeok}/XSN+\textit{i}/VCP+\textit{n}/ETM & world~class-\textsc{rel}\\
의상 &\textit{uisang} &&\textit{uisang}/NNG & fashion\\
디자이너 &\textit{dijaineo} && \textit{dijaineo}/NNG & designer\\
엠마누엘 &\textit{emmanuel} && \textit{emmanuel}/NNP & Emanuel\\
웅가로가 &\textit{unggaro-ga} && \textit{unggaro}/NNP+\textit{ga}/JKS& Ungaro-\textsc{nom}\\
실내 &\textit{silnae} &&\textit{silnae}/NNG& interior\\
장식용 &\textit{jangsikyong} &&\textit{jangsikyong}/NNG& decoration\\
직물 &\textit{jikmul} &&\textit{jikmul}/NNG& textile\\
디자이너로 &\textit{dijaineo-ro} && \textit{dijaineo}/NNG+\textit{ro}/JKB& designer-\textsc{ajt}\\
나섰다.& \textit{naseo-eoss-da.} & &\textit{naseo}/VV+\textit{eoss}/EP+\textit{da}/EF+./SF& become-\textsc{past}-\textsc{decl} \\
\end{tabular}}
\end{center}
\caption{
Morphological analysis and part of speech (POS) tagging example in the Sejong corpus: NN* are nouns, JK* are case makers and postpositions, V* are verbs, and E* are verbal endings.\label{ungaro-pos}}
\end{figure}

Every approach for Korean language processing has decided how to separate \textit{sequences} of morphemes into component parts, ranging from eojeols, a basic word-like unit, all the way down to a complete morphological parse. These decisions have been, for the most part, argued as either linguistically or technically motivated, with little or no interest in exploring an alternative. The choice does have some impact on the performance of algorithms in various tasks, such as part of speech (POS) tagging, syntactic parsing and machine translation. In the study, we analyze different granularity levels previously proposed and utilized for Korean language processing. In accordance with these analyzing works, we present the results of language processing applications using different segmentation granularity levels for future reference. To the best of the authors' knowledge, this is the first time that different granularity levels in Korean have been compared and evaluated against each other. This would contribute to fully understanding the current status of various granularity levels that have been developed for Korean language processing. Specifically the main goal of this paper is to diagnose the current state of natural language processing in Korean by tracing its development procedures and classifying them into five steps. Additionally, this paper aims to clearly explicate and evaluate the challenges unique to Korean language processing, with the objective of contributing to the improvement of various methodologies in this field. 
To this end, after presenting previous work in Section~\ref{previous-work}, the study introduces the segmentation granularity in Korean by classifying them into five different levels with a linguistic perspective as well as a natural language processing perspective in Section~\ref{definition}, and presents several application for Korean language processing using the five segmentation granularity levels by comparing them each other in Section~\ref{applications}. Finally, Section~\ref{conclusion} concludes the discussion.

\section{Previous work} \label{previous-work}

\begin{figure}
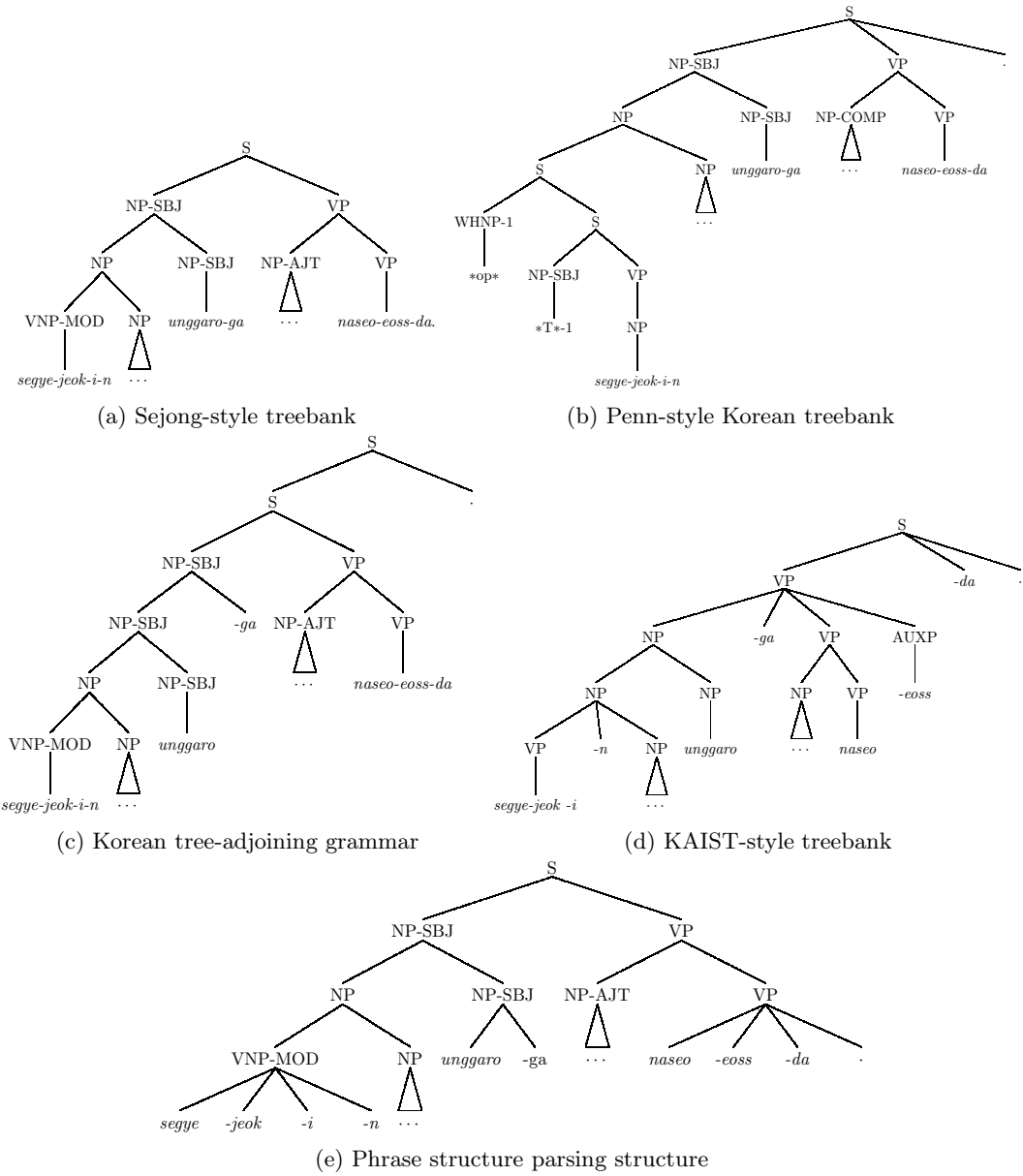

\centering

\begin{subfigure}[b]{0.39\textwidth}
\resizebox{\textwidth}{!}{
\synttree 
[S
	[NP-SBJ
	[NP
	[VNP-MOD [\textit{segye-jeok-i-n}] ]
[NP [.x $\cdots$] ]]
		[NP-SBJ [\textit{unggaro-ga}]] ]
	[VP
		[NP-AJT[.x $\cdots$ ]]
		[VP [ \textit{naseo-eoss-da$.$} ]]]]
}
\caption{Sejong{-style} treebank} 
\label{sejong-treebank}
\end{subfigure}
\begin{subfigure}[b]{0.51\textwidth}
\resizebox{\textwidth}{!}{
\synttree 
[S[NP-SBJ[NP [S [WHNP-1 [$*$op$*$]] [S [NP-SBJ [$*$T$*$-1]] [VP [NP [\textit{segye-jeok-i-n}]]]]] [NP [.x $\cdots$ ]] ][NP-SBJ [\textit{unggaro-ga}]]] [VP[NP-COMP[.x $\cdots$ ]] [VP [\textit{naseo-eoss-da}]]] [ $\cdot$ ]]
}
\caption{Penn{-style} Korean treebank} 
\label{penn-treebank}
\end{subfigure}

\begin{subfigure}[b]{0.44\textwidth}
\resizebox{\textwidth}{!}{
\synttree 
[S [S [NP-SBJ [NP-SBJ[NP[VNP-MOD [\textit{segye-jeok-i-n}]] [NP [.x $\cdots$ ]] ][NP-SBJ [\textit{unggaro}]] ] [\textit{-ga}]] [VP[NP-AJT[.x $\cdots$ ]] [VP [\textit{naseo-eoss-da}]]] ] [ $\cdot$ ]]
}
\caption{Korean tree-adjoining grammar}
\label{korean-ltag}
\end{subfigure}
\begin{subfigure}[b]{0.49\textwidth}
\resizebox{\textwidth}{!}{
\synttree 
[S[VP [NP[NP [VP [\textit{segye-jeok -i}] ] [\textit{-n}] [NP [.x $\cdots$ ]] ][NP [\textit{unggaro}] ]] [\textit{-ga}] [VP[NP[.x $\cdots$ ]] [VP [\textit{naseo}] ] ] [AUXP [\textit{-eoss}] ]] [ \textit{-da} ][ $\cdot$ ]]
}
\caption{KAIST{-style} treebank}
\label{kaist-treebank}
\end{subfigure}

\begin{subfigure}[b]{0.65\textwidth}
\resizebox{\textwidth}{!}{
\synttree 
[S[NP-SBJ[NP[VNP-MOD [\textit{segye}] [\textit{-jeok}] [\textit{-i}] [\textit{-n}]] [NP [.x $\cdots$ ] ] ][NP-SBJ [\textit{unggaro}] [-ga]]] [VP[NP-AJT[.x $\cdots$ ] ] [VP [\textit{naseo}] [\textit{-eoss}] [\textit{-da}] [ $\cdot$]]]]
}
\caption{Phrase structure parsing {structure}}
\label{phrase-structure-parsing-treebank}
\end{subfigure}

\caption{{Different} syntactic analyses using different segmentation granularities. POS labels are omitted. } 
\label{syntactic-granularity}
\end{figure}

Different granularity levels have been proposed mainly due to varying different syntactic analyses in several previously proposed Korean treebank datasets: KAIST \citep{choi-EtAl:1994}, Penn \citep{han-EtAl:2002}, and Sejong. While segmentation granularity, which we deal with, is based on morphological analysis, the syntactic theories are implicitly presented in the corpus for Korean words. 
Figure~\ref{syntactic-granularity} summarizes the syntactic trees which can be represented in Korean treebanks for different segmentation granularity levels. Korean TAG grammars \citep{park:2006} and CCG grammars \citep{kang:2011} described Korean by separating case markers. Most work on language processing applications such as phrase structure parsing and machine translation for Korean which uses the sentence by separating all morphemes \citep{choi-park-choi:2012:SP-SEM-MRL,park-hong-cha:2016:PACLIC,kim-park:2022}. The Penn Korean treebank introduced a tokenization scheme for Korean, while the KAIST treebank separates functional morphemes such as postpositions and verbal endings. Note that there are no functional tags (\textit{i.e.}, \texttt{-sbj} or \texttt{-ajt}) in the KAIST treebank.

Syllable-based granularity (\textit{e.g.}, \textit{se $\sqcup$ gye $\sqcup$ jeok $\sqcup$ i}, `world-class') \citep{yu-EtAl:2017:SCLeM,choi-EtAl:2017:SCLeM} and even character-based granularity using the Korean alphabet (\textit{s $\sqcup$ e $\sqcup$ g $\sqcup$ ye $\sqcup$ j $\sqcup$ eo $\sqcup$ k $\sqcup$ i}) \citep{stratos:2017:EMNLP,song-park:2020:TALLIP} have also been proposed  where $\sqcup$ indicates a blank space. They incorporate sub-word information to alleviate data sparsity especially in neural models. Dealing with sub-word level granularity using syllables and characters does not consider our linguistic intuition. We describe granularity based on a linguistically motivated approach in this paper, in which each segmentation is a meaningful morphological unit.

\section{Definition of segmentation granularity} \label{definition}

The annotation guidelines for Universal Dependencies stipulate each syntactic word, which is an atom of syntactic analysis, as a basic unit of dependency annotation \citep{nivre-EtAl:2020:LREC}. This stipulation presupposes that there must be a separate autonomous level of syntax and morphology. One of the features of agglutinative languages is that there is a one-to-one mapping between suffixes and syntactic categories, indicating that each suffix must have an appropriate syntactic category to which it belongs. More specifically, nouns can have individual markers indicating case, number, possessive, etc., whose orders are fixed. Thus, we can regard any given noun or verb as a stem followed by several inflectional or derivational morphemes. The number of slots for a given part of a category may be pretty high. In addition, an agglutinating language adds information such as negation, passive voice, past tense, honorific degree to the verb form. That is, in an agglutinating language, verbal morphemes are added to the root of the verb to convey various grammatical features, such as negation, passive voice, past tense, and honorific degree. One of the characteristics in Korean is to use a system of honorifics to express the hierarchical status and familiarity of participants in conversation with respect to the subject, object or the interlocutor. This system plays a great role in Korean grammar. When a speaker uses honorific expression, we can figure out the social relationship between the speaker, the interlocutor, and the object in the subject position at the same time. This honorific system is reflected in the honorific markers attached to the nouns, and verbal endings to the verb.

Such a complex and rich morphological system in agglutinative languages poses numerous challenges for natural language processing. The key obstacle lies in a voluminous number of word forms that can be derived from a single stem. Word form analysis involves breaking a given surface word form into a sequence of morphemes in an order that is admissible in Korean. However, several difficulties may arise in dividing these sequences of morphemes into appropriate units. This paper describes the segmentation granularity procedures that could influence the performance of algorithms in various tasks and the various analyses that have been adopted in Korean. First of all, we define five different levels of segmentation granularity for Korean, which have been independently proposed in previous work as different segmentation units. While Levels 1 (eojeols as they are), 2 (tokenization – separating words and symbols process), and 5 (separating all morphemes process) are due to technical reasons, Levels 3 (separating case markers process) and 4 (separating verbal endings process) are based on linguistic intuition.

\subsection{Level 1: eojeols} \label{l1}

As described previously, most Korean language processing systems and corpora have used the eojeol as a fundamental unit of analysis. For example, the Sejong corpus, the most widely-used corpus for Korean, uses the eojeol as the basic unit of analysis.\footnote{The Ministry of Culture and Tourism in Korea launched the 21st Century Sejong Project in 1998 to promote Korean language information processing.
The project has its name from Sejong the Great who conceived and led the invention of \textit{hangul}, the writing system of the Korean language.} The Sejong corpus was first released in 2003 and was continually updated until 2011. The project produced the largest corpus for the Korean language. It includes several types of corpora: historical, contemporary, and parallel text. Contents of the Sejong corpus represent a variety of sources: newswire data and magazine articles on various subjects and topics, several book excerpts, and scraped texts from the Internet. The Sejong corpus consists of the morphologically (part of speech tagged), the syntactically (treebank), and the lexical-semantically annotated text as well as a list of Korean words as dictionaries based on part of speech categories. Figure~\ref{sejong-corpus} shows an example of the Sejong corpus for the sentence in \eqref{ungaro}.

\begin{figure}
\begin{subfigure}[b]{\textwidth}
{\small
\begin{tabular}{lll}
BTAA0001-00000012 &프랑스의& 프랑스/NNP+의/JKG \\
BTAA0001-00000013 &세계적인& 세계/NNG+적/XSN+이/VCP+ㄴ/ETM \\
BTAA0001-00000014 &의상& 의상/NNG \\
BTAA0001-00000015 &디자이너& 디자이너/NNG \\
BTAA0001-00000016 &엠마누엘& 엠마누엘/NNP \\
BTAA0001-00000017 &웅가로가& 웅가로/NNP+가/JKS \\
BTAA0001-00000018 &실내& 실내/NNG \\
BTAA0001-00000019 &장식용& 장식/NNG+용/XSN \\
BTAA0001-00000020 &직물& 직물/NNG \\
BTAA0001-00000021 &디자이너로& 디자이너/NNG+로/JKB \\
BTAA0001-00000022 &나섰다.& 나서/VV+었/EP+다/EF+./SF \\
\end{tabular}
}
\caption{Morphologically (part of speech tagged) analyzed corpus where the word is analyzed as in the surface form. Therefore, even the punctuation mark is a part of the word.} 
\label{sejong-morph-corpus}
\end{subfigure}

\begin{subfigure}[b]{\textwidth}
{\small
\begin{tabular}{lll llll}
\multicolumn{7}{l}{}\\
(S &(NP-SBJ &(NP &\multicolumn{4}{l}{(NP-MOD 프랑스/NNP+의/JKG)}\\
&&&(NP &\multicolumn{3}{l}{(VNP-MOD 세계/NNG+적/XSN+이/VCP+ㄴ/ETM)}\\
&&&&(NP &\multicolumn{2}{l}{ (NP 의상/NNG) }\\
&&&&& (NP 디자이너/NNG)))) \\
&&(NP-SBJ &\multicolumn{4}{l}{(NP 엠마누엘/NNP)} \\
&&&\multicolumn{3}{l}{(NP-SBJ 웅가로/NNP+가/JKS)))} \\
& (VP& (NP-AJT& (NP& (NP&\multicolumn{2}{l}{(NP 실내/NNG)} \\
& &&&& \multicolumn{2}{l}{(NP 장식/NNG+용/XSN))} \\
& &&& \multicolumn{3}{l}{(NP 직물/NNG)) }\\
& && \multicolumn{4}{l}{(NP-AJT 디자이너/NNG+로/JKB))}\\
& & \multicolumn{5}{l}{(VP 나서/VV+었/EP+다/EF+./SF)))}\\
\end{tabular}
}
\caption{Syntactically analyzed corpus (treebank) where it inherits the annotation from the morphologically (part of speech tagged) analyzed corpus and added bracketing syntactic tree structure.} 
\label{sejong-treebank-figure1}
\end{subfigure}

\begin{subfigure}[b]{\textwidth}
{\small 
\begin{tabular}{lll}
\multicolumn{3}{l}{}\\
BSAA0001-00000012 &프랑스의& 프랑스/NNP+의/JKG \\
BSAA0001-00000013 &세계적인& 세계\_\_02/NNG+적/XSN+이/VCP+ㄴ/ETM \\
BSAA0001-00000014 &의상& 의상\_\_01/NNG \\
BSAA0001-00000015 &디자이너& 디자이너/NNG \\
BSAA0001-00000016 &엠마누엘& 엠마누엘/NNP \\
BSAA0001-00000017 &웅가로가& 웅가로/NNP+가/JKS \\
BSAA0001-00000018 &실내& 실내/NNG \\
BSAA0001-00000019 &장식용&  장식\_\_05/NNG+용/XSN \\
BSAA0001-00000020 &직물& 직물/NNG \\
BSAA0001-00000021 &디자이너로& 디자이너/NNG+로/JKB \\
BSAA0001-00000022 &나섰다.& 나서/VV+었/EP+다/EF+./SF \\
\end{tabular}
}
\caption{Semantically analyzed corpus where it includes the lexical semantic annotation to disambiguate the sense of the word by the Sejong dictionary.} 
\label{sejong-semantics}
\end{subfigure}

\caption{Example of the Sejong corpus for the sentence in \eqref{ungaro}}
\label{sejong-corpus}
\end{figure}

We define eojeols, as in the Sejong corpus, as granularity Level 1. Rationale of this segmentation granularity in Korean language processing is simply to use the word as it is in the surface form, in which the word is separated by a blank space in the sentence (that is, in a manner of what you see is what you get). Most morphological analysis systems have been developed based on eojeols (Level 1) as input and can yield morphologically analyzed results, in which a single eojeol can contain several morphemes. The dependency parsing systems described in \citet{oh-cha:2013} and \citet{park-EtAl:2013:IWPT} used eojeols as an input token to represent dependency relationships between eojeols. 
\citet{oh-EtAl:2011} {presented a system which predict phrase-level syntactic label for eojeols based on the sequence of morphemes in the eojeol.}
What is the most interesting is that \citet{petrov-das-mcdonald:2012:LREC} proposed Universal POS tags for Korean based on the eojeol and \citet{stratos-collins-hsu:2016:TACL} worked on POS tagging accordingly. Taking these basic trends into consideration, the study defines eojoeols as Level 1. Recently released KLUE (Korean Language Understanding Evaluation) also used the eojeol as a fundamental unit of analysis \citep{park-EtAl:2021:KLUE}.\footnote{\url{https://klue-benchmark.com}}

\subsection{Level 2: separating words and symbols} \label{l2}

The process of tokenization in the Korean language has often been overlooked, primarily because eojeols has traditionally been used as the basic unit of analysis. However, it has come to our attention that certain corpora have started adopting an English-like tokenization approach, which results in preprocessed words within these corpora. For example, the Penn Korean treebank \citep{han-EtAl:2002}, which punctuation marks are separated from words.\footnote{While the Penn Korean treebank separates all punctuation marks, quotation marks are the only symbols that are separated from words in the Sejong treebank {to distinguish between the quoted clause and the main sentence in the tree structure.}
We also note that among the existing corpora for Korean, only the Sejong treebank separates quotation marks from the word. 
Other Sejong corpora including the morphologically analyzed corpus do not separate the quotation marks, and still use the eojeol as a basic analysis unit.} 
This segmentation granularity especially in the Penn-treebank style corpus focuses on multilingual processing where Penn treebanks include English \citep{marcus-santorini-marcinkiewicz:1993:CL,taylor-marcus-santorini:2003}, Chinese \citep{xue-EtAl:2005}, Arabic \citep{maamouri-bies:2004} and Korean \citep{han-EtAl:2002}. The Penn Korean treebank follows the tokenization scheme that has been used in the other language of the Penn treebanks, as shown in Figure~\ref{penn-figure2}. The most distinctive feature in Level 2 lies in that the punctuation mark is all separated from the original word (tokenized).

\begin{figure} 
{\small
\begin{tabular}{llll llll}
(S& (NP-SBJ& \multicolumn{6}{l}{그/NPN+은/PAU)} \\
  & (VP& (S-COMP& (NP-SBJ& \multicolumn{3}{l}{르노/NPR+이/PCA)} \\
&&&(VP& (VP& (NP-ADV& \multicolumn{2}{l}{3/NNU}\\
&&&&&& \multicolumn{2}{l}{월/NNX+말/NNX+까지/PAU)}\\
&&&&& (VP& (NP-OBJ& \multicolumn{1}{l}{인수/NNC+제의/NNC}\\
&&&&&    &        & 시한/NNC+을/PCA)\\
&& & & & & \multicolumn{2}{l}{갖/VV+고/ECS))} \\
&&&&\multicolumn{4}{l}{있/VX+다/EFN+고/PAD))} \\
&&\multicolumn{6}{l}{덧붙이/VV+었/EPF+다/EFN)}\\
&\multicolumn{6}{l}{./SFN)}\\
\end{tabular}
}
\caption{Example of the Penn Korean treebank where the punctuation mark is separated from the word (tokenized): N* are nouns PA* are case markers and postpositions, V* are verbs, and E* are verbal endings.}
\label{penn-figure2}
\end{figure}

We define the tokenization by separating words and symbols as a granularity Level 2. 
\citet{chung-gildea:2009:EMNLP} used a granularity Level 2 for a baseline tokenization system for a machine translation system from Korean into English where they proposed an unsupervised tokenization method to improve the machine translation result. Figure~\ref{penn-figure2} illustrates that the punctuation marker has been separated from the verb \textit{deosbut-i-eoss-da} (`added') and assigned its own category with the marker being designated as `texttt{sfn}' in Penn Treebank. In addition, the tokenization schema of the sentence follows the method similar to the English language. That is, syntactic unit \textit{3-wol-mal-kka-ji} (`until the end of March') {is traditionally treated} as one eojeol, but in Level 2, this unit is tokenized as three different units such as \textit{3-wol} (`March'), \textit{mal} (end) and \textit{kka-ji} (`until'), which is tokenized identically to that of English such as until the end of March. As mentioned in Level 1, most Korean language processing systems have used an eojeol as their basic unit of analysis, resulting in a single eojeol involved with several different morphemes, which is a prominent feature in Level 1. According to this principle, we can easily identify that the noun phrase in a subject position \textit{geu-eun} forms one eojeol consisting of a stem \textit{geu} and a {topic} marker \textit{eun}. In the same way, the verb phrase \textit{deosbut-i-eoss-da} (`added') creates one eojeol with a root \textit{deosbut}, a {passive suffix} \textit{-i}, a past tense marker \textit{-eoss} and verb ending marker \textit{-da}.

\citet{park-EtAl:2014:ISWCPOSTER} also used this granularity to develop Korean FrameNet lexicon units by using the cross-lingual projection method from the Korean translation of the English Propbank \citep{palmer-gildea-kingsbury:2005:CL}. Universal Dependencies \citep{nivre-EtAl:2016:LREC,nivre-EtAl:2020:LREC} contains two Korean dependency treebanks, namely the GSD treebank \citep{mcdonald-EtAl:2013:ACL} and the KAIST treebank \citep{choi-EtAl:1994,chun-EtAl:2018:LREC}, which also use the tokenization scheme by separating words and punctuation marks.

{Recently, \citet{park-kim:2023} insisted that the functional morphemes in Korean should be treated as part of a word in Korean categorial grammars, with the result that their categories for detailed morphemes do not require to be assigned individually in a syntactic level, and also that it would be more efficient to assign the syntactic categories on the fully inflected lexical word derived by the lexical rule of the morphological processes in the lexicon.}

\subsection{Level 3: separating case markers} \label{l3}

From a purely linguistic perspective, postpositions as functional morphemes in Korean convey grammatical cases (\textit{e.g.}, nominative or accusative), adverbial relations (spatial, temporal or directional), semantic roles and conjunctives by combining with the lexical words. We may separately indicate them as case marker, adverbial postposition, auxiliary postposition, and conjunctive postposition, respectively, though we generally term them as postpositions or case markers, depending on the authors. In linguistics, a marker also refers to a free or bound morpheme indicating the grammatical function of the word, phrase or sentence. For the sake of convenience, the paper uses case markers as a term for covering them. Case markers are immediately attached following a noun or pronoun. They are used to indicate the grammatical roles of a noun in a sentence such as subject, object, complement or topic.

First of all, \textit{-i} and \textit{-ga} are nominative case markers whose form depends on whether the stem ends with a vowel or consonant. When the honorific subject is used, this nominative case marker will be replaced by the honorific marker \textit{-kkeseo}, instead of \textit{-i} or \textit{-ga}. An honorific is marked to encode the relative social status of the interlocutors. A major feature of this honorific system is typically to convey the same message in both honorific and familiar forms. Korean honorifics are added to nouns, verbs, and adjectives. Similarly to this nominative case marker, the honorific dative case marker \textit{-kke} will be used instead of the familiar dative case marker \textit{-ege}. The rest of the markers are used to express the adverbial relations such as directional, temporal, spatial including source and destination, and accompaniment.
All of these markers attached to the noun stem cannot be duplicated, showing {complementary distribution}. As shown in the example \eqref{holangi-ga}, the nominative case marker \textit{-ga} cannot be together with the instrumental case marker \textit{-ro} in \eqref{holangi-ga-ro}, and cannot collocate also with the dative case marker \textit{-ege} in \eqref{holangi-ga-ege}.

\begin{exe} 
\ex \label{holangi-ga}
    \begin{xlist}
        \ex \gll {\textit{holangi-ga}} {\textit{sanab-da}}. \\
            {tiger-\textsc{nom}}  {fierce-\textsc{decl}}  \\
            \trans 'A tiger is fierce.'        
        \ex \label{holangi-ga-ro}  
            \gll *\textit{holangi-ga-ro} \textit{sanab-da}.\\
            {tiger-\textsc{nom}-\textsc{\textit{`to'}}}  {fierce-\textsc{decl}}  \\
            \trans 'A tiger is fierce.'
        \ex \label{holangi-ga-ege} 
            \gll *\textit{holangi-ga-ege} \textit{sanab-da}.\\
            {tiger-\textsc{nom}-\textsc{dat}}  {fierce-\textsc{decl}}  \\
            \trans 'A tiger is fierce.'      
    \end{xlist}
\end{exe}

Under a perspective of natural language processing, the Sejong corpus has been criticized for the scope of the case marker, in which only a final noun (usually the lexical anchor) in the noun phrase is a modifier of the case marker. For example, \textit{Emmanuel Ungaro-ga} in the Sejong corpus is annotated as (NP (NP \textit{Emmanuel}) (NP \textit{Ungaro-ga})), in which only \textit{Ungaro} is a modifier of \textit{-ga} (`\textsc{nom}'). 
For example as described in \citet{ko:2010}, while there are several debates on whether a noun or a case marker is a modifier in Korean, this is beyond the scope of the paper. The Penn Korean treebank does not explicitly represent this phenomenon. 
It just groups a noun phrase together: \textit{e.g.}, (NP \textit{Emmanuel Ungaro-ga}), which seems to be treated superficially as a simple compound noun phrase. Collins' preprocessing for parsing the Penn treebank adds intermediate NP terminals for the noun phrase \citep{collins:1997:ACL,bikel:2004:CL}, and so NPs in the Penn Korean treebank will have a similar NP structure to the Sejong corpus \citep{chung-post-gildea:2010:SPMRL}. 
To fix the problem in the previous treebank annotation scheme, there are other annotation schemes in the corpus and lexicalized grammars. They are introduced to correctly represent the scope of the case marker. \citet{park:2006} considered case markers as independent elements within the formalism of Tree adjoining grammars \citep{joshi-levy-takahashi:1975}. Therefore, he defined case markers as an auxiliary tree to be adjoined to a noun phrase. In contrast to case markers, verbal endings in the inflected forms of predicates are still in the part of the eojoel and they are represented as initial trees for Korean TAG grammars. The lemma of the predicate and its verbal endings are dealt with as inflected forms instead of separating functional morphemes \citep{park:2006}. 
This idea is going back to Maurice Gross's lexicon grammars in 1970s \citep{gross:1975} and his students who worked on a descriptive analysis of Korean in which the number of predicates in Korean could be fixed by generating all possible inflection forms: \textit{e.g.}, \citet{pak:1987,nho:1992,nam:1994,shin:1994,park:1996,chung:1998,han:2000}.

\subsection{Level 4: separating verbal endings} \label{l4}

With a purely linguistic perspective, Korean verbs are formed in terms of the agglutinating process by adding various endings to the stem. Korean is widely known to have a great many verbal endings between this stem and final verbal endings. More specifically, the verbal endings in Korean are well known to be complex in their syntactic structures in the sense that the verbal endings carry much of functional load in the grammatical aspects such as sentence mood, tense, voice, aspect, honorific, conjunction, etc.: 
for example, 
\textit{inter alia}, 
{tense \citep{byungsun:2003}, 
grammatical voice \citep{chulwoo:2007}, 
interaction of tense–aspect–mood marking with modality \citep{jaemog:1998}, 
evidentiality \citep{donghoon:2008}, and 
interrogativity \citep{donghoon:2011}.}
More additional endings can be used to denote various semantic connotations. That is, a huge number of grammatical functions are achieved by adding various verbal endings to verbs. 
The number can also vary depending on the theoretical analyses, naturally differing in their functions and meanings. These endings, of course, do not change the argument structures of a predicate. A finite verb in Korean can have up to seven suffixes as its endings, whose order is fixed. As mentioned in the previous section, the Korean honorific system can also be reflected in verbs with honorific forms. When a speaker expresses his respect toward the entities in a subject or indirect object position, the honorific marker \textit{-(eu)si} is attached to the stem verb, thereby resulting in the verb form \textit{sanchaegha} (`take a walk'). The suffixes denoting tense, aspect, modal, formal, mood are followed by the honorific.

Unlike the markers attached to nouns, {Korean verbal endings are added to the verb stem in a specific order, depending on the tense, mood, and politeness level of the sentence,} as illustrated in \eqref{halabeoji}. The verb stem \textit{sanchaegha} (`to take a walk') can be followed by the honorific \textit{-si} in \eqref{halabeoji-sanchaegha-si-n-da}. 
The two suffixes indicating an honorific and past tense can be attached to the verb stem in \eqref{halabeoji-sanchaegha-sy-eoss-da}. 
One more additional suffix of retrospective aspect is added in the example in \eqref{halabeoji-sanchaegha-sy-eoss-deon-jeog-i-iss-da}. 
If the order of a past suffix and honorific suffix is changed in the verbal endings, the sentence would be ungrammatical, as in \eqref{halabeoji-kkeseo-sanchaegha-eoss-sy-deon-jeog-i-iss-da}.

\begin{exe}
\ex \label{halabeoji}
\begin{xlist}
\ex \label{halabeoji-sanchaegha-si-n-da}
\gll \textit{halabeoji-kkeseo}  \textit{jamsi} \textit{sanchaegha-si-n-da}.\\
{Grandfather-\textsc{nom-hon}} {for~a~while}  {take~a~walk-\textsc{hon}-\textsc{pres}-\textsc{decl}}   \\
\trans 'Grandfather takes a walk for a moment.'

\ex \label{halabeoji-sanchaegha-sy-eoss-da}  
\gll \textit{halabeoji-kkeseo} \textit{jamsi} \textit{sanchaegha-sy-eoss-da}.\\
{Grandfather-\textsc{hon}} {for a while} {take~a~walk-\textsc{hon}-\textsc{past}-\textsc{decl}}\\
\trans 'Grandfather took a walk for a moment.'

\ex \label{halabeoji-sanchaegha-sy-eoss-deon-jeog-i-iss-da}
\gll \textit{halabeoji-kkeseo} \textit{jamsi} \textit{sanchaegha-sy-eoss-deon} \textit{jeog-i} \textit{iss-da}.\\
Grandfather-\textsc{hon} {once} {go~for~a~walk-\textsc{hon}-\textsc{past}-\textsc{asp}} {experience-\textsc{cop}} {be-\textsc{decl}} \\
\trans 'Grandfather once went for a walk for a moment.'

\ex  \label{halabeoji-kkeseo-sanchaegha-eoss-sy-deon-jeog-i-iss-da}
\gll *\textit{halabeoji-kkeseo} \textit{jamsi} \textit{sanchaegha-eoss-sy-deon} \textit{jeog-i} \textit{iss-da}.\\
Grandfather-\textsc{hon} {once} {go~for~a~walk-\textsc{past}-\textsc{hon}-\textsc{asp}}  {experience-\textsc{cop}} {be-\textsc{decl}}   \\
\trans 'Grandfather once went for a walk for a moment.'
\end{xlist}
\end{exe}

\begin{figure}
    \centering
\begin{forest}
[CP [C\texttt{'} [IP   [NP [$\cdots$] ] [I\texttt{'} [VP [$\cdots$] 
[V  [\textit{naseo}] ]] [I [T [\textit{-eoss}] ]]]] [C [\textit{-da}] ]]]
\end{forest}    
    \caption{{GB theory for Korean syntactic analyses, in which the entire sentence depends on verbal endings}}
    \label{kaist-ex}
\end{figure}
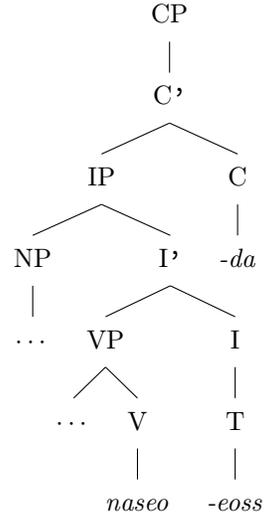

Government and Binding (GB) theory \citep{chomsky:1981,chomsky:1982} for Korean syntactic {analyses}, in which the entire sentence depends on verbal endings as described in Figure \ref{kaist-ex} for \textit{naseo-eoss-da} (`became'). This means that the functional morpheme \textit{-eoss} is assigned its own syntactic category T(ense) and the verbal ending \textit{-da} C(omplimentizer) attached in the final position determines the whole syntactic category CP in Korean.

From the Natural Language Processing perspective, the KAIST treebank \citep{choi-EtAl:1994}, an earliest Korean treebank, introduced this type of analysis, which is Level 4. It is the granularity Level 4 that we adapt the KAIST treebank representation. While the KAIST treebank separates case markers and verbal endings with their lexical morphemes, punctuation marks are not separated and they are still a part of preceding morphemes as represented in the Sejong treebank. Therefore, strictly speaking, one could judge that the KAIST treebank is not granularity Level 4 by our definition because we separate punctuation marks. In addition, while it also represents derivational morphology in the treebank annotation (\textit{i.e.}, for a copula \textit{segye-jeok $\sqcup$ -i $\sqcup$ -n} (`world-class') in the KAIST treebank), we separate only verbal endings (\textit{i.e.}, \textit{segye-jeok-i $\sqcup$ -n}).

\subsection{Level 5: separating all morphemes} \label{l5}

Many downstream applications for Korean language processing are based on the granularity Level 5, in which all morphemes are separated: 
POS tagging \citep{jung-lee-hwang:2018:TALLIP,park-tyers:2019:LAW}, 
phrase-structure parsing \citep{choi-park-choi:2012:SP-SEM-MRL,park-hong-cha:2016:PACLIC,kim-park:2022} and 
statistical machine translation (SMT) \citep{park-hong-cha:2016:PACLIC,park-EtAl:2017:Cupral}, etc. where the applications take all the morphemes separated sequence instead of the surface sentence segmented by a blank, as input for language processing. 
{A morpheme-based annotation scheme proposed in \citet{park-tyers:2019:LAW} for POS tagging has been extended to dependency parsing \citep{chen-EtAl:2022:COLING} and named-entity recognition \citep{chen-lim-park-2023-ner} and it attained the most advanced evaluation outcomes.}
Figure \ref{level5-all-morphemes} shows examples of the downstream application process: constituent parsing using the Sejong treebank and machine translation from Korean into English. The sentence often in these applications is converted into the sequence of morphemes to be parsed or translated. They mostly implement granularity level 5 to avoid the problems of data sparsity and unknown words because the number of possible types combined in longer segmentation granularities, such as eojeol, can increase exponentially. Such morpheme-based analysis for the word can be generated by a morphological analysis system. Therefore, most POS tagging systems can produce segmentation granularity Level 5. Separating these morphemes is straightforward from such morphological analysis results. 
{For instance, 
as shown in Figure~\ref{level5-all-morphemes}, in Level 5, 
The phrase \textit{segyejeokin} (`world-class'), which also includes derivational morphemes, is treated as a separated four morphemes sequence \textit{segye-jeok-i-n} instead of one surface segment as input for language processing. 
Specifically, this phrase is assigned four different categories: a NNG (common noun for \textit{segye}), XSN (nominal derivational affix for \textit{jeok}), VCP (copular for \textit{i}) and ETM (adnominal affix for \textit{n}), respectively. These categories consist of the word stem, two derivational morphemes, and an inflectional morpheme, resulting in a new category verb functioning as a modifier in this sentence.}

\begin{figure} 
\begin{subfigure}[b]{\textwidth}
\resizebox{\textwidth}{!}{
\synttree 
[SENT [S [NP-SBJ [NP [NP-MOD 
[NNP [프랑스] ] 
[JKG [의]]]
[NP [VNP-MOD 
[NNG [세계] ] 
[XSN [적] ] 
[VCP [이] ] 
[ETM [ㄴ]
]][NP [NP 
[NNG [의상] ]
][NP 
[NNG [디자이너] ] 
]]]][NP-SBJ [NP 
[NNP [엠마누엘] ]
][NP-SBJ 
[NNP [웅가로] ] 
[JKS [가] ] 
]]][VP [NP-AJT [NP [NP [NP 
[NNG [실내] ]
][NP 
[NNG [장식] ] 
[XSN [용] ] 
]][NP 
[NNG [직물] ]
]][NP-AJT 
[NNG [디자이너] ] 
[JKB [로] ] 
]][VP 
[VV [나서] ] 
[EP [었] ] 
[EF [다] ] 
[SF [$\cdot$] ]
]]] ]
}
\caption{Morpheme-based treebank for constituent parsing \citep{choi-park-choi:2012:SP-SEM-MRL}} \label{sjtree-constituent-converted}
\end{subfigure}

\begin{subfigure}{\textwidth}
{\small
\begin{tabular}{c}
\\
{프랑스의~세계적인~의상~디자이너~엠마누엘~웅가로가~실내~장식용~직물~디자이너로~나섰다. } \\
{} \\
$\Downarrow$ {\footnotesize (separating all morphemes)} \\
{} \\
{프랑스~의~세계~적~이~ㄴ~의상~디자이너~엠마누엘~웅가로~가~실내~장식~용~직물~디자이너~로~나서~었~다~. } \\
{} \\$\Downarrow$ {\footnotesize (machine translation)}\\
{} \\
\textit{The world-class French fashion designer Emanuel Ungaro became an interior textile designer.} \\
\end{tabular}
}
\caption{Example of a machine translation process \citep{park-EtAl:2017:Cupral}} \label{mt-converted}
\end{subfigure}

\caption{Example of downstream application processes} \label{level5-all-morphemes}
\end{figure}

\subsection{Discussion} \label{granularity-discussion}

\begin{figure}
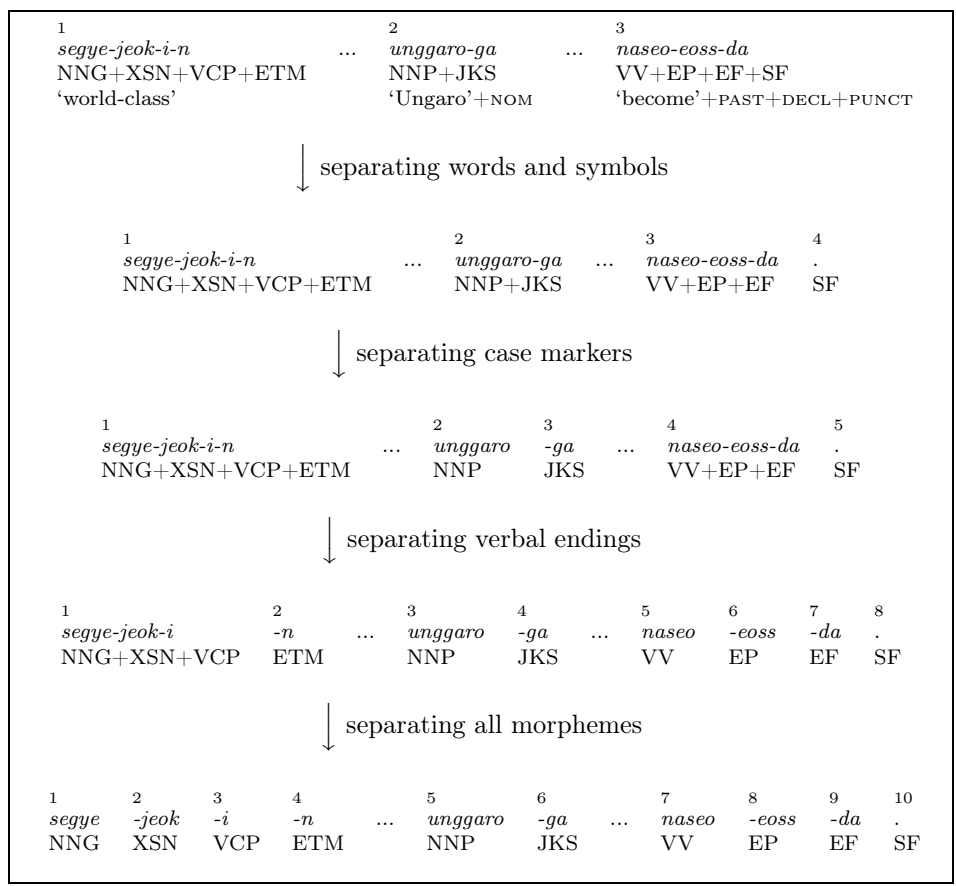

\centering
\begin{tabular} {|c|} \hline
\begin{footnotesize}
\begin{tabular} {lllll} 
$_1$ & & $_2$ && $_3$ \\
\textit{segye-jeok-i-n} &...  & \textit{unggaro-ga} & ... & \textit{naseo-eoss-da} \\
NNG+XSN+VCP+ETM & & NNP+JKS && VV+EP+EF+SF \\
`world-class' & & `Ungaro'+\textsc{nom} && `become'+\textsc{past}+\textsc{decl}+\textsc{punct} \\
\end{tabular}\end{footnotesize} \\~\\ 
$\Big\downarrow$ separating  words and symbols \\~\\

\begin{footnotesize}
\begin{tabular} {llll ll }
$_1$ & & $_2$ & & $_3$ & $_4$\\
\textit{segye-jeok-i-n} &... &\textit{unggaro-ga} &... & \textit{naseo-eoss-da} & . \\
NNG+XSN+VCP+ETM & & NNP+JKS && VV+EP+EF & SF \\
\end{tabular} \end{footnotesize} \\~\\ $\Big\downarrow$ separating case markers \\~\\

\begin{footnotesize}
\begin{tabular} {llll lll}
$_1$ & &$_2$ & $_3$ && $_4$ & $_5$\\
\textit{segye-jeok-i-n} &...& \textit{unggaro} & \textit{-ga} &... & \textit{naseo-eoss-da} & . \\
NNG+XSN+VCP+ETM & & NNP & JKS && VV+EP+EF & SF \\
\end{tabular} \end{footnotesize} \\~\\ $\Big\downarrow$ separating verbal endings \\~\\

\begin{footnotesize}
\begin{tabular} {lllll lllll}
$_1$ & $_2$ & & $_3$ & $_4$ && $_5$ & $_6$ &$_7$&$_8$\\
\textit{segye-jeok-i} & \textit{-n} & ...&\textit{unggaro} & \textit{-ga} &... & \textit{naseo} & \textit{-eoss} & \textit{-da} & .  \\
NNG+XSN+VCP & ETM & & NNP & JKS && VV & EP & EF & SF \\
\end{tabular} \end{footnotesize} \\~\\ $\Big\downarrow$ separating all morphemes \\~\\

\begin{footnotesize}
\begin{tabular} {lllll lll llll}
$_1$ & $_2$ & $_3$ & $_4$ & & $_5$ & $_6$ & &$_7$&$_8$& $_9$& $_{10}$ \\
\textit{segye} &\textit{-jeok} &\textit{-i} & \textit{-n} &...& \textit{unggaro} & \textit{-ga} & {...} & \textit{naseo} & \textit{-eoss} & \textit{-da} & .  \\
NNG & XSN & VCP & ETM & & NNP & JKS && VV & EP & EF & SF \\
\\
\end{tabular}\end{footnotesize} 
\\\hline
\end{tabular}
\caption{Five levels of segmentation granularity in Korean and their POS annotation.} \label{granularity}
\end{figure}

Figure~\ref{granularity} summarizes an example of each segmentation granularity level. For our explanatory purpose, we use the following sentence in \eqref{ungaro}: \textit{segye-jeok-i-n ... unggaro-ga ... naseo-eoss-da.} (`The world-class ... Ungaro became ...'). The advantage of Level 1 is that it has many linguistics resources to work with. The main weakness of Level 1 is that it requires segmentation including the tokenization process which has been a main problem in language processing in Korean.
While Level 2 has appeared more frequently especially in recent Universal Dependencies (UD)-related resources, and Levels 3 and 4 propose an analysis more linguistically pertinent, they do not mitigate the segmentation problem. Level 5 has the practical merits of a processing aspect. However, the eventual problem for the reunion of segmentation morphemes, for example the generation task in machine translation, still remains, and it has not been discussed much yet.

\section{Diagnostic analysis} \label{applications} 
In this section, we present several applications for Korean language processing using proposed segmentation granularity levels to compare them to each other. We use the default options that the system provides for experiments. For experiments, we convert all data sets into each segmentation granularity. We utilize a 90-10 split for the Sejong treebank for the training and evaluation for POS tagging and syntactic parsing. We utilize training and evaluation data sets for Korean-English machine translation provided by \citet{park-hong-cha:2016:PACLIC}.

Firstly, Table~\ref{stat} shows the number of tokens, the ratio of morphologically complex words (MCW) which are made up of two or more morphemes, and the number of immediate non-terminal (NT) nodes (the number of monomorphemic and complex word patterns) in the entire Sejong treebank. Therefore, the immediate NT nodes signify the POS labels, and can be eojeols, morphemes and symbols according to different segmentation granularity.

\begin{table}
\centering
\begin{tabular} {r  ccccc l} 
\hline
& Level 1 &Level 2 & Level 3 & Level 4 & Level 5 & \\ 
\noalign{\smallskip}\hline\noalign{\smallskip} 
Token & 370,729 & 436,448 & 577,153 & 752,654 & 829,506 & \\
MCW & 0.7881 & 0.6451 & 0.2939 & 0.0934 & 0 & \\
Immediate NT &4,318&2,378&1,228&526&45 & \\ \noalign{\smallskip} \hline
\end{tabular}
\caption{
The number of tokens, the ratio of the morphologically complex words (MCW) and the number of immediate non-terminal (NT) in the corpus} \label{stat} 
\end{table}

\subsection{Language processing tasks} \label{lang-tasks}

\paragraph{Word segmentation, morphological analysis and POS tagging}
Word segmentation, morphological analysis and {POS tagging} for Korean requires detection of morpheme boundaries. We use UDPipe, a trainable pipeline \citep{straka-hajic-strakova:2016:LREC} to perform tokenizing and POS tagging tasks. The current experimental setting achieved the {state} of the art word segmentation and POS tagging result for Korean \citep{park-tyers:2019:LAW}. 
Each trained POS tagging model assigns POS labels for its tokens of granularity. For example, a model should generate \textit{segye+jeok+i+n} for morpheme boundary and \texttt{nng+xsn+vcp+etm} as a single POS label in Level 1 for \textit{segyejeokin} (`world-class'), or \texttt{nng} in Level 5 for \textit{segye} (`world').  
{We present the f1 score ($2 \cdot \frac{\text{precision} \cdot \text{recall}}{\text{precision} + \text{recall}}$) for word segmentation evaluation using precision and recall described in \eqref{word-segmentation-metric}, and} the accuracy score for POS tagging evaluation as in \eqref{pos-metric}.

{
\begin{align}
\begin{split}\label{word-segmentation-metric}
\text{precision} &= \frac{\text{\# of relevant word segments} \cap \text{\# of retrieved word segments}}{\text{\# of retrieved word segments}}\\
\text{recall} &   = \frac{\text{\# of relevant word segments} \cap \text{\# of retrieved word segments}}{\text{\# of relevant word segments}}
\end{split}
\end{align}
}

{
\begin{align}
\begin{split}\label{pos-metric}
\text{accuracy} &= \frac{\text{correct \# of POS tagging labels}}{\text{total \# of POS tagging labels}}
\end{split}
\end{align}
}

\paragraph{Syntactic parsing}
Using the granularity Level 5 has been the de facto standard for Korean phrase structure parsing \citep{choi-park-choi:2012:SP-SEM-MRL,park-hong-cha:2016:PACLIC,kim-park:2022}. 
We train and evaluate the Berkeley parser \citep{petrov-EtAl:2006:COLACL,petrov-klein:2007:main} with the different granularity levels. 
The Berkeley parser uses the probabilistic CFG with latent annotations previously proposed in \citet{matsuzaki-miyao-tsujii:2005:ACL}, and performs a series of split and merge cycles of non-terminal nodes to maximize the likelihood of a treebank. It still shows relatively good parsing results. We keep the structure of the Sejong treebank, and terminal nodes and their immediate NTs are varied depending on the granularity level. We provide gold POS labels as input instead of predicting them during parsing to the original word boundary in the word. This allows us to evaluate parsing results with the same number of terminals for all granularity levels. We present the f1 score by precision and recall of bracketing using EVALB \citep{black-etal-1991-procedure} for parsing evaluation {which uses the f1 score based on precision and recall presented in \eqref{parsing-metric}}.

{
\begin{align}
\begin{split}\label{parsing-metric}
\text{precision} &= \frac{\text{\# of relevant constituents} \cap \text{\# of retrieved constituents}}{\text{\# of retrieved constituents}}\\
\text{recall} &   = \frac{\text{\# of relevant constituents} \cap \text{\# of retrieved constituents}}{\text{\# of relevant constituents}}
\end{split}
\end{align}
}

\paragraph{Machine translation}

Using the granularity Level 5 has been the de facto standard for machine translation for Korean \citep{park-hong-cha:2016:PACLIC,park-EtAl:2017:Cupral}. We use the Moses statistical machine translation system \citep{koehn-EtAl:2007:PosterDemo} with the different granularity levels for Korean to train the phrase-based translation model and minimum error rate training \citep{och:2003:ACL} during validation. We present the BLEU {(BiLingual Evaluation Understudy)} score \citep{papineni-EtAl:2002:ACL} for evaluation.

\subsection{Results and discussion}

\begin{table}
\centering
\begin{tabular} {r  ccccc l} 
\hline
& Level 1 &Level 2 & Level 3 & Level 4 & Level 5 & \\ 
\noalign{\smallskip}\hline\noalign{\smallskip} 
Segmentation & \textbf{100.00} & 95.43 & 94.31 & 93.05 & 90.15 & (\textsc{f$_1$}) \\
POS tagging  &83.18&86.28&89.21&92.82& \textbf{96.01}& (\textsc{acc})\\
Syntactic parsing  &76.69&77.50&81.54&\textbf{84.64}&82.23& (\textsc{f$_1$}) \\
Machine translation &5.86&6.87&7.64&7.85& \textbf{7.98}& (\textsc{bleu})\\ \noalign{\smallskip} \hline
\end{tabular}
\caption{
Experiment results on POS tagging, syntactic parsing and machine translation based on different segmentation granularity levels. For comparison purposes, we convert POS tagging results into Level 1 and syntactic parsing results into Level 5. Translation direction is Korean into English.
} \label{results} 
\end{table}

The direct interpretation of task results between the different granularity levels would be difficult because the levels of representation are different (\textit{e.g.}, the number of lexical tokens is different in Table~\ref{stat}). For comparison purposes of experiment results, 
(1) we report segmentation results where Level 1 does not require any segmentation. 
(2) We convert all POS tagging results into Level 1 based on eojeol after training and predicting results for each segmentation granularity level. Therefore, the presented POS tagging accuracy is based on Level 1 eojeols as in previous work on POS tagging \citep{cha-EtAl:1998,hong:2009,na:2015:TALLIP}. 
(3) We convert syntactic parsing results into morpheme-based Level 5 as in previous work on phrase structure parsing \citep{choi-park-choi:2012:SP-SEM-MRL,park-hong-cha:2016:PACLIC,kim-park:2022}. Although the Berkeley parser can predict the POS label during parsing, we provide gold POS labels, {which is correct POS labels from the test dataset} as input for the parsing system to keep original morpheme boundaries. After parsing sentences for each segmentation granularity level, we convert parsing results into Level 5. 
(4) For machine translation, we translate Korean sentences in different segmentation granularity into English where there is no different segmentation granularity. {We uses \texttt{multi-bleu.perl} provided by Moses \citep{koehn-EtAl:2007:PosterDemo} to evaluate the translation result.\footnote{\url{https://github.com/moses-smt/mosesdecoder}}}

All results based on different segmentation granularity levels are reported in Table~\ref{results}. 
The interpretation of results of segmentation is straightforward where no tokenization is required in Level 1 and more tokenization is required in Level 5. From POS tagging to MT, we provide a gold-segmented sequence to evaluate each task. Results of POS tagging indicate that such morpheme-based analysis outperform other granularity, which conforms to the previous results on morphological analysis and POS tagging for Korean \citep{park-tyers:2019:LAW}. 
As we described, fine-grained granularity by separating all morphemes (Level 5) has been utilized for downstream applications such as machine translation for Korean and it shows the best performance in the BLEU score \citep{papineni-EtAl:2002:ACL}. 
Whereas phrase structure parsing also uses by separating all morphemes (Level 5) as input for the previous parsing system \citep{choi-park-choi:2012:SP-SEM-MRL,park-hong-cha:2016:PACLIC,kim-park:2022}, granularity by separating only functional morphemes including case markers and verbal endings and keeping other affixes for morphological derivation (Level 4) outperform Level 5. 
The modern statistical parsers have used markovization annotation for non-terminal nodes to elaborate context-free grammar rules for parsing either using the manual heuristics \citep{johnson:1998:CL,klein-manning:2003:ACL} or machine learning techniques \citep{petrov-EtAl:2006:COLACL,petrov-klein:2007:main}. Parsing performance in the statistical parsers is directly related with the size and the quality of CFG rules generated by these annotation schemes of non-terminal nodes. The other explanation for Level 4's parsing performance involves its linguistically soundness of its segmentation of the word, in which its immediate non-terminal nodes represent actual part-of-speech information of the word with its adjoined functional morphemes. Linguistic information of this kind might help to improve the representation of the treebank grammar that is implied by the parsing system.

\section{Conclusion} \label{conclusion}

The study addresses word segmentation granularity  for the segmentation in Korean language processing. There have been multiple possible word segmentation granularity levels from a word to morphemes in Korean, and for specific language processing and annotation tasks, several different granularity levels have been proposed and developed. It is that the agglutinative languages including Korean can have a one-to-one mapping between functional morpheme and syntactic category, even though the annotation guidelines for Universal Dependencies typically regard a basic unit of dependency annotation as a syntactic word. We have presented five different levels of segmentation granularity in Korean. We have analyzed and compared these levels of granularity by using Korean language applications as well. Previous work for Korean language processing has not explicitly mentioned which level of segmentation granularity is used, and this makes it difficult to properly compare results between systems. As described, these different levels of segmentation granularity could exist mainly because various Korean treebanks represent their syntactic structure differently. These treebanks also use the different segmentation of words depending on their linguistic and computational requirements. While a certain segmentation granularity may be well suited for some linguistic phenomena or applications, we need to find a correct segmentation granularity level to adapt to our requirements and expectations for Korean language processing.

\section*{Acknowledgement}
The work has started when Jungyeul Park was at University of Arizona during 2016-2017. 
The authors thank Mike Hammond, Francis Tyers and Shane Steinert-Threlkeld for their discussion on the earlier version of this manuscript, and Eric VanLieshout for proofreading.
The authors also thank anonymous reviewers who have generously provided valuable feedback.


\appendix

\section{Experiment details} \label{experiment-details}
This section details the configuration of experiments including their datasets. 

\subsection{Word segmentation, morphological analysis and POS tagging}
We use UDPipe \citep{straka-hajic-strakova:2016:LREC} to perform tokenizing (word segmentation) and POS tagging tasks based on the experiment setting described by \citet{park-tyers:2019:LAW}. It achieves the state of the art word segmentation and POS tagging result for Korean, and provides the detailed procedure for for word segmentation and and POS tagging using UDPipe.\footnote{\url{https://zenodo.org/record/3236528}}
Figure~\ref{pos-experiments} shows examples of the CoNLL-U style dataset for word segmentation, morphological analysis and POS tagging for each segmentation granularity levels.
For experiments, we convert all sentences in the Sejong treebank into each segmentation granularity with their morphological analysis. We utilize a 90-10 split for the training and evaluation based on \citet{kim-park:2022}, which proposed the standard split for the Sejong treebank.

\begin{figure}
\centering
\scriptsize
\begin{subfigure}[b]{\textwidth}
\centering
{
\begin{tabular} {lll ll } 
\multicolumn{5}{l}{...}\\
2&   세계적인&  세계+적+이+ㄴ&  NOUN & NNG+XSN+VCP+ETM\\
\multicolumn{5}{l}{...}\\
6&   웅가로가&  웅가로+가&  PROPN  & NNP+JKS  \\
\multicolumn{5}{l}{...}\\
11&  나섰다.  &나서+었+다+.  &VERB&  VV+EP+EF+SF   \\
\end{tabular} 
}
\caption{Level 1} 
\label{pos-level1}
\end{subfigure}

\begin{subfigure}[b]{\textwidth}
\centering
{
\begin{tabular} {lll ll }
\multicolumn{5}{l}{}\\
\multicolumn{5}{l}{...}\\
2&   세계적인&  세계+적+이+ㄴ&  NOUN & NNG+XSN+VCP+ETM\\
\multicolumn{5}{l}{...}\\
6&   웅가로가&  웅가로+가&  PROPN  & NNP+JKS  \\
\multicolumn{5}{l}{...}\\
11&  나섰다  &나서+었+다  &VERB&  VV+EP+EF  \\
12&  .  &.  &PUNCT &  SF   \\
\end{tabular} 
}
\caption{Level 2} 
\label{pos-level2}
\end{subfigure}

\begin{subfigure}[b]{\textwidth}
\centering
{
\begin{tabular} {lll ll } 
\multicolumn{5}{l}{}\\
\multicolumn{5}{l}{...}\\
3&   세계적인&  세계+적+이+ㄴ&  NOUN & NNG+XSN+VCP+ETM\\
\multicolumn{5}{l}{...}\\
7-8&웅가로가&\_&\_\\
7&웅가로&웅가로&PROPN&NNP\\
8&가&가&ADP&JKS\\
\multicolumn{5}{l}{...}\\
14&  나섰다  &나서+었+다  &VERB&  VV+EP+EF  \\
15&  .  &.  &PUNCT &  SF   \\
\end{tabular} 
}
\caption{Level 3} 
\label{pos-level3}
\end{subfigure}

\begin{subfigure}[b]{.58\textwidth}
\centering
{
\begin{tabular} {lll ll }
\multicolumn{5}{l}{}\\
\multicolumn{5}{l}{...}\\
3-4&세계적인&\_&\_&\\
3&세계적이&세계+적+이&VERB&NNG+XSN+VCP \\
4&ㄴ&은&PART&ETM\\
\multicolumn{5}{l}{...}\\
8-9&웅가로가&\_&\_\\
8&웅가로&웅가로&PROPN&NNP\\
9&가&가&ADP&JKS\\
\multicolumn{5}{l}{...}\\
15-17&나섰다&\_&\_\\
15&나서&나서&VERB&VV\\
16&었&었&PART&EP\\
17&다&다&PART&EF\\
18&.&.&PUNCT&SF\\
\end{tabular} 
}
\caption{Level 4} 
\label{pos-level4}
\end{subfigure}
\begin{subfigure}[b]{.38\textwidth}
\centering
\begin{tabular} {lll ll } 
\multicolumn{5}{l}{}\\
\multicolumn{5}{l}{...}\\
3-6&세계적인&\_&\_&\\
3&세계&세계&NOUN&NNG \\
4&적&적&PART&XSN\\
5&이&이&VERB&VCP\\
6&ㄴ&은&PART&ETM\\
\multicolumn{5}{l}{...}\\
10-11&웅가로가&\_&\_\\
10&웅가로&웅가로&PROPN&NNP\\
11&가&가&ADP&JKS\\
\multicolumn{5}{l}{...}\\
18-20&나섰다&\_&\_\\
18&나서&나서&VERB&VV\\
19&었&었&PART&EP\\
20&다&다&PART&EF\\
21&.&.&PUNCT&SF\\
\end{tabular} 
\caption{Level 5} 
\label{pos-level5}
\end{subfigure}

\caption{{Examples of the CoNLL-U style dataset for word segmentation, morphological analysis and POS tagging}}
\label{pos-experiments}
\end{figure}

\begin{figure}
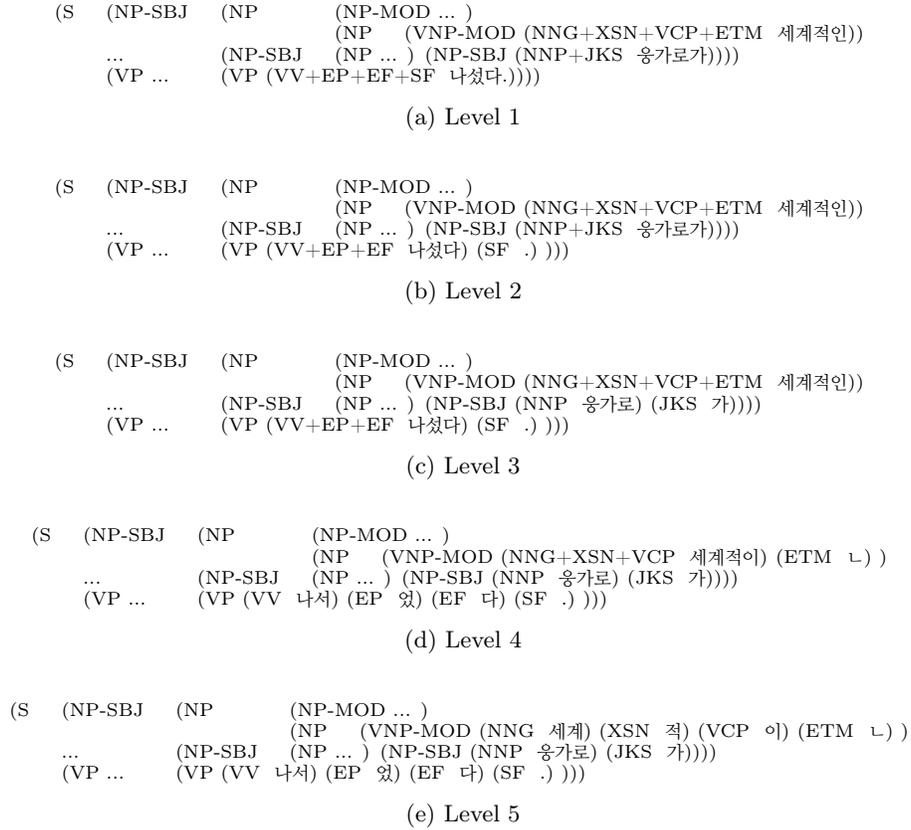

\centering
\scriptsize
\begin{subfigure}[b]{\textwidth}
\centering
{
\begin{tabular}{lll llll}
(S &(NP-SBJ &(NP &\multicolumn{4}{l}{(NP-MOD ... )}\\
&&&(NP &\multicolumn{3}{l}{(VNP-MOD (NNG+XSN+VCP+ETM~ 세계적인))}\\
&...      & (NP-SBJ &\multicolumn{4}{l}{(NP ... ) (NP-SBJ (NNP+JKS~ 웅가로가))))} \\
& (VP ... & \multicolumn{5}{l}{(VP (VV+EP+EF+SF~ 나섰다.))))}\\
\end{tabular}
}
\caption{Level 1} 
\label{parsing-level1}
\end{subfigure}

\begin{subfigure}[b]{\textwidth}
\centering
{
\begin{tabular}{lll llll}
\multicolumn{7}{l}{}\\
\multicolumn{7}{l}{}\\
(S &(NP-SBJ &(NP &\multicolumn{4}{l}{(NP-MOD ... )}\\
&&&(NP &\multicolumn{3}{l}{(VNP-MOD (NNG+XSN+VCP+ETM~ 세계적인))}\\
&...      & (NP-SBJ &\multicolumn{4}{l}{(NP ... ) (NP-SBJ (NNP+JKS~ 웅가로가))))} \\
& (VP ... & \multicolumn{5}{l}{(VP (VV+EP+EF~ 나섰다)  (SF~ .) )))}\\
\end{tabular}
}
\caption{Level 2} 
\label{parsing-level2}
\end{subfigure}

\begin{subfigure}[b]{\textwidth}
\centering
{
\begin{tabular}{lll llll}
\multicolumn{7}{l}{}\\
\multicolumn{7}{l}{}\\
(S &(NP-SBJ &(NP &\multicolumn{4}{l}{(NP-MOD ... )}\\
&&&(NP &\multicolumn{3}{l}{(VNP-MOD (NNG+XSN+VCP+ETM~ 세계적인))}\\
&... & (NP-SBJ &\multicolumn{4}{l}{(NP ... ) (NP-SBJ (NNP~ 웅가로) (JKS~ 가))))} \\
& (VP ... & \multicolumn{5}{l}{(VP (VV+EP+EF~ 나섰다)  (SF~ .) )))}\\

\end{tabular}
}
\caption{Level 3} 
\label{parsing-level3}
\end{subfigure}

\begin{subfigure}[b]{\textwidth}
\centering
{
\begin{tabular}{lll llll}
\multicolumn{7}{l}{}\\
\multicolumn{7}{l}{}\\
(S &(NP-SBJ &(NP &\multicolumn{4}{l}{(NP-MOD ... )}\\
&&&(NP &\multicolumn{3}{l}{(VNP-MOD (NNG+XSN+VCP~ 세계적이) (ETM~ ㄴ)  )}\\
&... & (NP-SBJ &\multicolumn{4}{l}{(NP ... ) (NP-SBJ (NNP~ 웅가로) (JKS~ 가))))} \\
& (VP ... & \multicolumn{5}{l}{(VP (VV~ 나서) (EP~ 었) (EF~ 다)  (SF~ .) )))}\\

\end{tabular}
}
\caption{Level 4} 
\label{parsing-level4}
\end{subfigure}

\begin{subfigure}[b]{\textwidth}
\centering
{
\begin{tabular}{lll llll}
\multicolumn{7}{l}{}\\
\multicolumn{7}{l}{}\\
(S &(NP-SBJ &(NP &\multicolumn{4}{l}{(NP-MOD ... )}\\
&&&(NP &\multicolumn{3}{l}{(VNP-MOD (NNG~ 세계)  (XSN~ 적) (VCP~ 이)  (ETM~ ㄴ)  )}\\
&... & (NP-SBJ &\multicolumn{4}{l}{(NP ... ) (NP-SBJ (NNP~ 웅가로) (JKS~ 가))))} \\
& (VP ... & \multicolumn{5}{l}{(VP (VV~ 나서) (EP~ 었) (EF~ 다)  (SF~ .) )))}\\
\end{tabular}
}
\caption{Level 5} 
\label{parsing-level5}
\end{subfigure}

\caption{{Examples of the Penn treebank style dataset for syntactic parsing}}
\label{parsing-experiments}
\end{figure}

{
\subsection{Syntactic parsing}
We train and evaluate the Berkeley parser \citep{petrov-EtAl:2006:COLACL,petrov-klein:2007:main} with the different granularity levels. 
For parsing experiments, we keep the original structure of the Sejong treebank, and only terminal nodes (`word') and their immediate non terminal nodes (`POS label') are varied depending on the granularity level.
Figure~\ref{parsing-experiments} shows examples of the Penn treebank style dataset for syntactic parsing. 
We followed training and  parsing procedures provided the Berkeley parser \citep{petrov-EtAl:2006:COLACL,petrov-klein:2007:main} using the proposed dataset.\footnote{\url{https://github.com/slavpetrov/berkeleyparser}}
For experiments, we convert all tree structures in the Sejong treebank into each segmentation granularity. We also utilize a 90-10 split for the training and evaluation based on \citet{kim-park:2022}.
}

{
\subsection{Machine translation}
We use the Moses statistical machine translation system \citep{koehn-EtAl:2007:PosterDemo} with the different granularity levels for Korean to train the phrase-based translation model and minimum error rate training \citep{och:2003:ACL} during validation. 
Figure~\ref{mt-experiments} shows examples of sentences for machine translation. 
We followed training and translation procedure provided by Moses using the proposed dataset.\footnote{\url{http://www2.statmt.org/moses/?n=Moses.Baseline}}
For experiments, we convert all sentences in Korean from the Korean-English parallel corpus\footnote{\url{https://github.com/jungyeul/korean-parallel-corpora}} into each segmentation granularity. 
We utilize a dataset split for the training and evaluation based on \citet{park-hong-cha:2016:PACLIC}.
}

\begin{figure}
    \centering
\footnotesize
\begin{tabular}{c  c}
Level 1     &  ... 세계적인 ... 웅가로가 ... 나섰다. \\
Level 2     &  ... 세계적인 ... 웅가로가 ... 나섰다~ .\\
Level 3     &  ... 세계적인 ... 웅가로~ 가 ... 나섰다~ .\\
Level 4     &  ... 세계적이~ ㄴ ... 웅가로~ 가 ... 나서~ 었~ 다~ .\\
Level 5     &  ... 세계~ 적~ 이~ ㄴ ... 웅가로~ 가 ... 나서~ 었~ 다~ .\\ 
\multicolumn{2}{c}{}\\
(translation) & \textit{The world-class ... Ungaro became ... .} \\
\end{tabular}    
    \caption{{Examples of sentences for Korean-English machine translation}}
    \label{mt-experiments}
\end{figure}

\end{document}